%% file: iclr2021_conference.tex
\documentclass{article} 
\usepackage{iclr2021_conference,times}

\input{math_commands.tex}

\usepackage{url}
\usepackage{multirow}
\usepackage{graphicx}
\usepackage{subfigure}
\usepackage{float}
\usepackage{natbib}

\title{Digging Deeper into CRNN Model in Chinese Text Images Recognition}

\iclrfinalcopy
\author{Kunhong Yu\textsuperscript{1}  \& Yuze Zhang \textsuperscript{2}  \\
\texttt{\textsuperscript{1}yuranusduke@163.com} \& \texttt{\textsuperscript{2}zyz9629@gmail.com}\\ 
}

\begin{document}

\maketitle

\begin{abstract}
Automatic text image recognition is a prevalent application in computer vision field. One efficient way is use Convolutional Recurrent Neural Network(CRNN) to accomplish task in an end-to-end(End2End) fashion. However, CRNN notoriously fails to detect multi-row images and excel-like images. In this paper, we present one alternative to first recognize single-row images, then extend the same architecture to recognize multi-row images with proposed multiple methods. To recognize excel-like images containing box lines, we propose Line-Deep Denoising Convolutional AutoEncoder(Line-DDeCAE) to recover box lines. Finally, we present one Knowledge Distillation(KD) method to compress original CRNN model without loss of generality. To carry out experiments, we first generate artificial samples from one Chinese novel book, then conduct various experiments to verify our methods.
\end{abstract}

\section{Introduction}

Optical Character Recognition(OCR) has pervasive applications in the real world. Current methods tend to use Deep Neural Network(DNN) to receive simply preprocessed image as input, then extract useful features to do final predictions (\cite{girshick2014rich, wang2012end, bissacco2013photoocr, jaderberg2016reading, graves2008novel}). This naive procedure has become the standard method of this task, more researches are now heavily emphasized on how to invent smart tricks to improve performance in the overall architecture.

Convolutional Recurrent Neural Network(CRNN) (\cite{shi2016end}) is the first successful end-to-end(End2End) model to recognize text images. Multiple new ideas were put forward to boost accuracy based on CRNN model subsequently. CRNN consists of three interdependent modules, Convolutional Neural Network(CNN), Recurrent Neural Network(RNN), Connectionist Temporal Classification(CTC). CNN receives image to extract features and RNN(in particular, Bidirectional LSTM/BiLSTM) loops over all features to calculate CTC loss. To update parameters in the model, CRNN backpropagates the whole model to obtain gradients.

CRNN behaves nicely when input image has only one row of texts. Intuitively, CNN utilizes larger kernel size in width than in height, therefore, extracted features can be input into BiLSTM as embeddings to compute CTC loss. In order to recognize multi-row images, existing methods tend to use object detection algorithms to firstly obtain regions of text, then put them into CRNN. This two-stage procedure works well when large training data set is available, however, model capacity and computation efficiency seriously limit deployment. 

Moreover, according to our experimental results, we conclude that CRNN performs less accurately when image takes on excel-like form. Existing methods utilize simple traditional computer vision algorithms to get rid of box lines in the image ahead. Nevertheless, this method is not robust to different image styles under various circumstances.

In this paper, we present multiple methods to overcome above difficulties in simple End2End CRNN model. To recognize single-row images, original CRNN applies VGG16 model (\cite{simonyan2014very}) as backbone with modified kernel size and stride in CNN block to derive $1*1$ feature map in the final \texttt{conv} layer. To avoid well-designed hyperparameters, we utilize standard VGG16 model(cut off a few layers according to sample complexity), in order to obtain final embeddings to input into BiLSTM, we design \textit{column average pooling} to adaptively get $1*1$ feature map.

To recognize multi-row images, we propose two simple yet effective approaches. First one is in the last layer of CNN, for each output feature map, we simply stretch it out into one row of features. All feature maps can be executed in parallel. According to our implementations, we surprisingly find out this method can recognize multi-row images nicely. Another one is we use attention mechanism in the last feature maps of CNN, for each feature map, we reshape it into two dimensional matrix with height in the first dimension and the product of width and number of channels in the second, then we use two independent \texttt{Conv1d} layers to \texttt{conv} over the first dimension to get new feature maps and attention masks, we refer to attention since we adopt the similar idea from ECANet (\cite{wang2020eca}).

Our next task is to eliminate box lines in the images if needed. We propose one Deep Denoising Convolutional AutoEncoder(DDeCAE) (\cite{gondara2016medical}) variant Line-DDeCAE to treat texts as noise trying to reconstruct box lines. Inspired by Denoising AutoEncoder, we view box lines lying in the lower manifold in high dimensional space of data set than texts, so it is at ease to recover them, finally, adding original input data gives us clean images without box lines.

To deploy our model into portable devices, we choose to compress large model into a light one. Different strategies have been established. Pruning (\cite{han2015learning, luo2017thinet, molchanov2016pruning, li2016pruning}) cuts abundant weights in the network and fine-tune the whole model over and over; Binarized Net (\cite{dean2012large, rastegari2016xnor, hubara2016binarized}) treats weights either -1 or 1 to reduce storage for the model, etc.. All above works tend to obtain small architecture via loss of learned information more or less. Knowledge Distillation(KD) (\cite{hinton2015distilling}) is one promising method to transfer learned knowledge from large(teacher) model to small(student) model through imitation of logits between those two. In order to efficiently shrink model size, we adopt KD methods in three aspects. First, we apply standard KD to match soft logits, then borrow the idea from FitNets (\cite{romero2014fitnets}) to match intermediate layers in CNN, finally, we also push hidden states and cell states in BiLSTM to be matched. 

To summarize, our contributions in this paper are : 

\qquad1. For single-row images, we propose to use simple \textit{column average pooling} rather than well-designed hyperparameters to obtain $1 * 1$ feature map of the last layer of CNN.

\qquad2. For multi-row images, we propose two simple yet effective methods to boost performance.

\qquad3. For images with box lines, we propose Line-DDeCAE to recover lines in the images, then we can obtain clean text images by summation with inputs.

\qquad4. To deploy our model into portable devices, we propose a KD method for CRNN which compresses large model into a lightweight one.

\qquad5. We generate one artificial Chinese data set for verification, extensive experiments reveal efficiency of our methods.

The paper is organized as follows: section \ref{rw} discusses related work with our methods; section \ref{methods} introduces proposed single-row and multi-row images recognition algorithms and delves deep into Line-DDeCAE model and elaborates KD method for CRNN compression; experiments are conducted in Section \ref{experiments}; final conclusion is drawn in section \ref{conclusion}.

\section{Related Work}
\label{rw}
We briefly review related works of CRNN, attention, AutoEncoder and KD.

\textbf{CRNN.} CRNN model has gained much attention in text images recognition. Being known for End2End training and inference fashion, CRNN discards complicated procedures for image preprocessing, text detection and text segmentation with final recognition. CRNN simply applies VGG16 as backbone to extract features from images to automatically accomplish text detection. To implement text segmentation and recognition, CRNN utilizes BiLSTM and CTC loss to wrap up whole process for predictions. CRNN is one of default architectures in research area nowadays. Some other CRNN-like models are proposed (\cite{kapka2019sound, kao2018r}). However, CRNN is not robust to multi-row images and excel-like images. To recognize multi-row images, one simple way is first employ object detection algorithms(PSENet, YOLO, etc.) (\cite{ li2018shape, redmon2016you, ren2015faster}) to detect regions of text, then extracted original images can be put into CRNN model for training, this procedure is hardly executed in End2End fashion, therefore, two-stage process can cut off connection between them. Moreover, excel-like images contain box lines, our experimental results demonstrate CRNN is not friendly with this kind of data. Finally, CRNN can't satisfy low latency implementation because of the usage of large model backbone.
In this paper, we put our heavy emphasis on baseline CRNN model to tackle with above problems with proposed multiple simple yet effective methods. 

\textbf{Attention.} Attention mechanism (\cite{bahdanau2014neural}) has seeped into broader areas in machine learning nowadays. In Natural Language Processing(NLP), attention has gained enough momentum. Recently, Transformer (\cite{vaswani2017attention}) model applies pure attention to beat traditional language models. BERT (\cite{devlin2018bert}) learns contextual word embeddings which has been proved effectively in many downstream NLP tasks. More and more NLP models surpass human-level performance also encapsulating attention, like GPT (\cite{khandelwal2019sample}), T5 (\cite{raffel2019exploring}), etc..
In Computer Vision(CV), attention was famously used in object recognition task (\cite{fu2017look}). Some other attention mechanisms are also available, SENet (\cite{hu2018squeeze}) first introduced channel-wise attention by computing sigmoidal attention value for each channel's globally average-pooled activations. More methods were put forward based on SENet's attention (\cite{wang2020eca, li2019selective}). In this paper, we apply the same idea from ECANet (\cite{wang2020eca}) where attention was replaced by \texttt{Conv1d} layer rather than fully connected layer in original SENet.

\textbf{AutoEncoder.} Inspired by PCA, AutoEncoder(AE) was proposed to be used to do pre-training for large neural network when large training data set was not available (\cite{ng2011sparse, gondara2016medical, rifai2011contractive}). AE is composed of encoder which is used to compress inputs into dense features and decoder which maps dense features back into inputs space hoping outputs could be similar to inputs. AE can be trained either in greedy layer-wise fashion where inputs are reconstructed layer by layer or in end-to-end fashion where all layers can be trained at the same time, namely deep AE or stacked AE. Moreover, AE can be viewed as generative model to generate new examples from compressed hidden features. Variational AE(VAE) (\cite{kingma2013auto}) firstly maps inputs into Gaussian distribution in the compressed layer, then maps sampled compressed(hidden) feature into various outputs. In this paper, we use one variant deep AE where encoder is made up of convolutional layers and pooling layers, decoder is made up of upsampling layers and transpose convolutional layers.

\textbf{Knowledge Distillation.} There are various approaches to implement model compression. Pruning (\cite{han2015learning, luo2017thinet, molchanov2016pruning, li2016pruning}) removes unimportant weights during training, and fine-tune the whole model over and over. Quantization (\cite{wu2016quantized, han2015deep}) aims to share the same value with weights which are in the similar range. Binarized Net (\cite{dean2012large, rastegari2016xnor, hubara2016binarized}) tries to satisfy all weights being -1 or 1. Moreover, some light model designs are also available, we refer readers to  (\cite{sandler2018mobilenetv2, iandola2016squeezenet, zhang2018shufflenet}). Knowledge Distillation(KD) (\cite{hinton2015distilling}) is one very promising method to transfer knowledge from large(teacher) model to small(student) model. Rather than directly learn from data 'hard' labels, standard KD methods propose student model learns both 'hard' labels and teacher's output 'soft' labels, which appear to be the softmax probability distribution with temperature hyperparameter. Plenty of KD algorithms have been published to tackle with shortcomings in standard KD. FitNets (\cite{romero2014fitnets}) was the first KD method trying to let student model mimic intermediate layers in the teacher model. In FSP (\cite{yim2017gift}), student learns Gram matrix of teacher across the large part of model. Attention KD (\cite{zagoruyko2016paying}) tries to learn attention values in intermediate layers between student and teacher. In channel-wise distillation, \cite{zhou2020channel} tries to match SE attentions between teacher and student model. Above works were all assuming there is a fixed pre-trained teacher model, therefore, one has to store the output of teacher model for downstream student training, which is not efficient. To address this, works to do online distillation emerge. DML (\cite{zhang2018deep}) tends to make teacher learn from student as well with standard KD method but they are all learning simultaneously. Be-your-own-teacher (\cite{zhang2019your}) learns student from itself architecture. \cite{chen2017learning} invented simple and effective object detection distillation method where teacher and student model are learning jointly. Now more works are heavily on how to learn both teacher and student at the same time. In this paper, we also adopt this idea to learn large CRNN and small CRNN models together without needing to store pre-trained teacher's statistics, which yields high efficiency.

\section{Methods}
\label{methods}
\subsection{Column Average Pooling}

\begin{figure}[h]
	\begin{center}
		\centering
		\includegraphics[width=9cm,height=4cm]{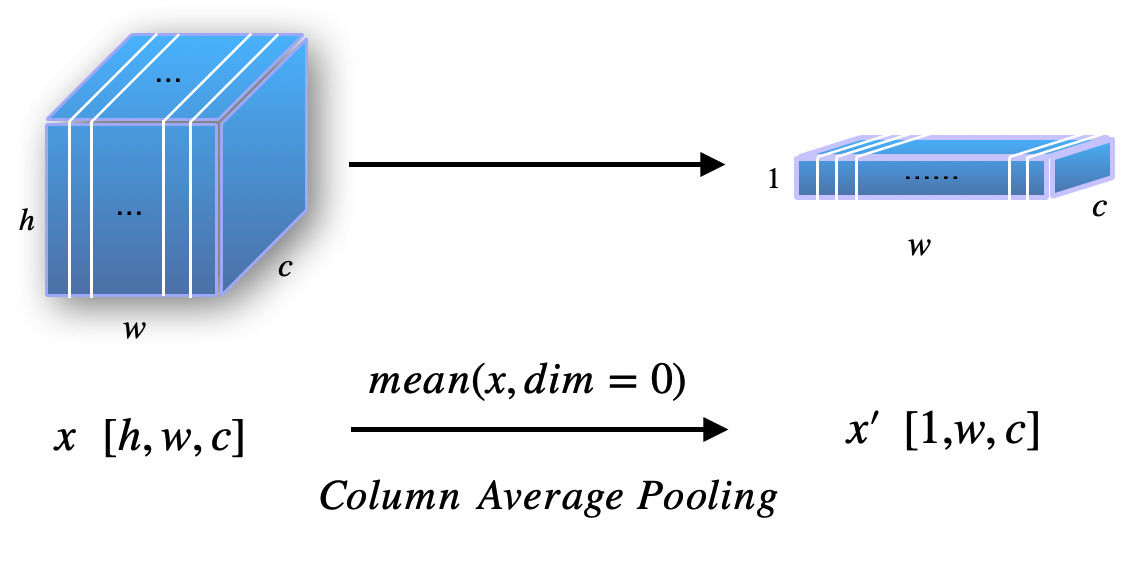}
	\end{center}
	\caption{Column Average Pooling}\label{cap_fig}
\end{figure}

Traditional CRNN redesigns the hyperparameters in CNN module to satisfy the size of $1 * 1$ of feature map of the last layer. However, we found Column Average Pooling(CAP) may lead to the similar result without careful hyperparameters tuning, which is illustrated in Figure~\ref{cap_fig}.

Suppose the size of feature map $x$ of the last layer has shape: $[h, w, c]$, where $h>1, w, c$ are the height, width and the number of channels respectively. Here, we omit the batch dimension. CAP is really simple, let $x'\in\mathbb{R}^{w\times c}$ be the output of CAP, then

\begin{equation}
	x'[i, j] = mean(x[:, i, j], dim = 0)
\end{equation}

\subsection{MultiRow Stretching}
\begin{figure}[h]
	\begin{center}
		\centering
		\includegraphics[width=9cm,height=4cm]{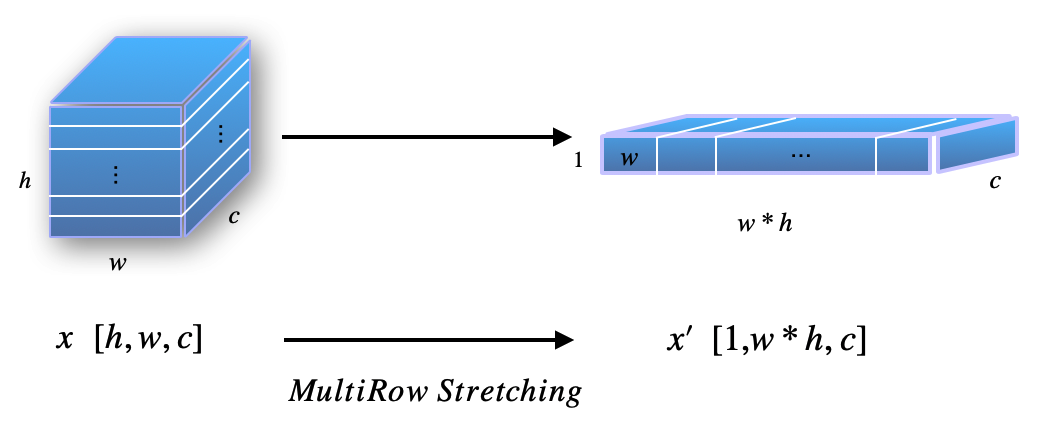}
	\end{center}
	\caption{MultiRow Stretching}\label{ms_fig}
\end{figure}

In order to tackle with multi-row images, we first use one simple method to stretch final layer's feature map into one row. Inspired by FCN in objection detection (\cite{dai2016r}), each sliding window can be processed in parallel using \texttt{conv} operation, so the final layer's feature map can be decomposed into different parts corresponding to each window. For multi-row images, each row can be operated in parallel as well, therefore, we adopt the same idea from FCN, using the same architecture for recognizing single-row images, we can derive multi-row feature map in the last layer illustrated in Figure \ref{ms_fig}, so we reshape it into a single row feature map in MultiRow Stretching method. 

Intuitively, we treat all sentences in one image as one sentence by concatenating them together in row manner. Specifically, reuse the notation from above section, to calculate CTC loss finally, we do the following operations in PyTorch fashion with label $y$:
\begin{align}
	x&=view(x, (w*h, c)) \\
	\mathcal{L}_{ctc}&=CTCLoss(BiLSTM(x), y)
\end{align}

\subsection{Attention}
\begin{figure}[h]
	\begin{center}
		\centering
		\includegraphics[width=14cm,height=7cm]{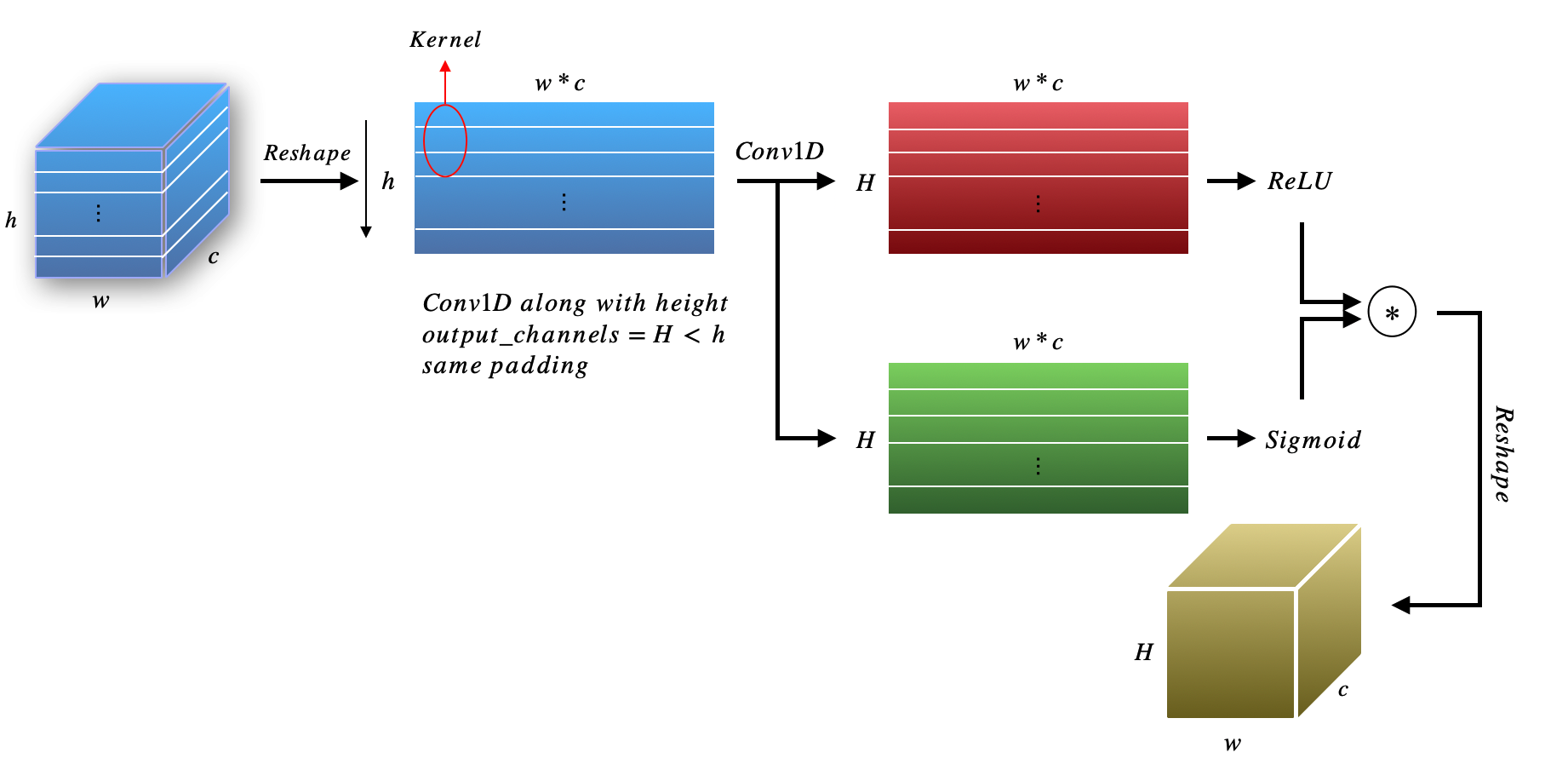}
	\end{center}
	\caption{Attention}\label{attention}
\end{figure}

MultiRow Stretching can tackle with images with few characters, the vanishing or exploding gradient problem may occur if the stretched features are long. Moreover, if there is no evident relationship among each row in the original input image, then this method may converge to sub-optimal result. Therefore, we use similar attention method from ECANet (\cite{wang2020eca}) to derive final feature map for image with fixed rows. Figure \ref{attention} illustrates this procedure.

Following the above notations, we first reshape 3D tensor $x\in\mathbb{R}^{h\times w\times c}$ into 2D tensor $x\in\mathbb{R}^{h\times (w\times c)}$. For each row of feature, we treat it as one element of final feature map. Intuitively, suppose the rows of each image is $H<h$, so we attend each row in $x$ to derive both feature map and sigmoidal attention mask. Like in ECANet, we use \texttt{Conv1D} to \texttt{conv} over the height of feature map. Final results can be derived by the product of attended feature maps and attention masks.
To summarize the above statement, we have the following operations:
 
\begin{align}
	\small
	x&=view(x, (h, w*c)) \\
	x'&=Conv1D(w*c, w*c, kernel\_size=h-H+1)(x) \\
	x'&=ReLU(x)\\
	attention&=Conv1D(w*c, w*c, kernel\_size=h-H+1)(x) \\
	mask&=sigmoid(attention) \\
	x&=x'*mask
\end{align}

For final $x$, we separate it into independent parts row by row. Then we input them into BiLSTM to calculate individual CTC loss. Given batch size $B$, the optimized CTC loss is :

\begin{equation}
 	\mathcal{L}_{ctc} = \frac{1}{BH}\sum_{i=1}^{B}\sum_{j=1}^{H}CTCLoss(x_{j}^{(i)}, y_{j}^{(i)})
\end{equation}

where $x_{j}^{(i)}$ is the $i^{th}$ example of $j^{th}$ part of feature map, $y_{j}^{(i)}$ is the corresponding label.

\subsection{Line-Deep Denoising Convolutional AutoEncoder}
\begin{figure}[t]
	\begin{center}
		\centering
		\includegraphics[width=13cm,height=4cm]{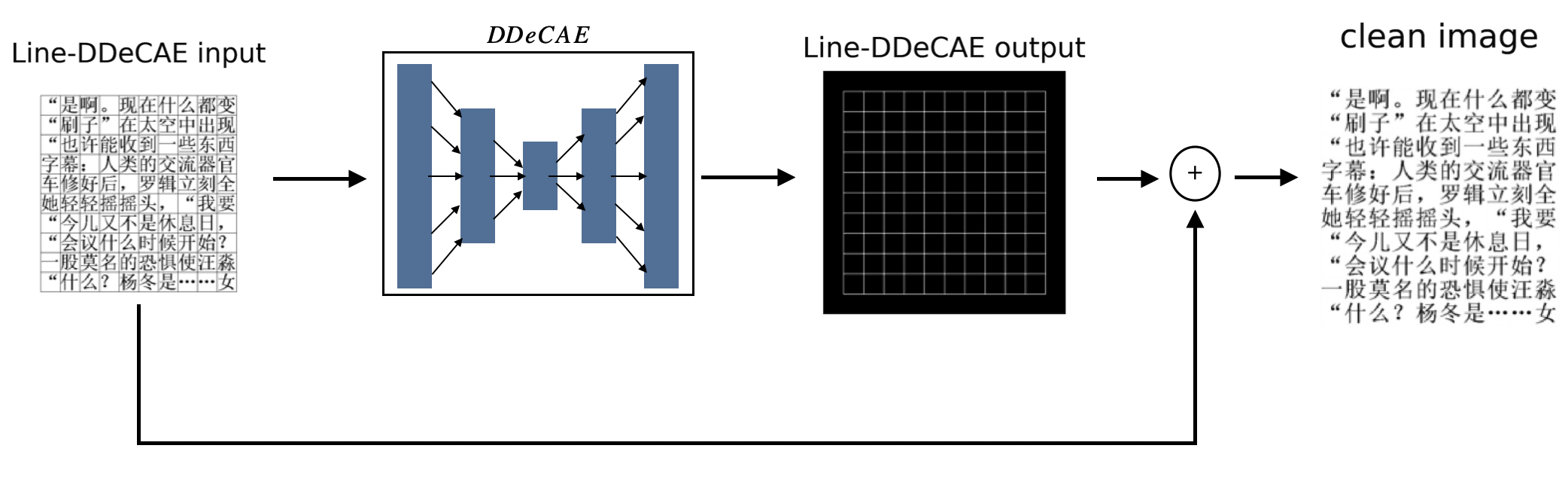}
	\end{center}
	\caption{Line-DDeCAE architecture}\label{ddecae}
\end{figure}

In this section, we propose one Deep Denoising Convolutional AutoEncoder(DDeCAE) (\cite{gondara2016medical}) variant, called Line-DDeCAE in Figure \ref{ddecae} to strip away box lines in the original images. Original images with excel-like form are harder to recognize. Therefore, in this work, we tend to first obtain clean images without box lines, then send them into CRNN to do recognition task.

Inspired by Denoising AutoEncoder(DeAE), the original input is distorted by a small random noise to be compressed into tight features and reconstruct the original input. Intuitively, adding noise lets original data distribution draft away a little, which lead to more robust compressed features. In this work, we either treat texts in the image as noise trying to recover box lines or box lines in the image as noise trying to recover texts(which is what DDeCAE does). We use the same architectures in DCGAN's  (\cite{radford2015unsupervised}) discriminator and generator to construct Line-DDeCAE's encoder and decoder respectively with few modifications. 

In experiments, we find when we treat box lines as noise and recover texts, the results are really noisy. Since texts lie in the more complicated manifold in high dimension than box lines, therefore stripping box lines is one easier task, recovering texts is harder. As a result, we try to recover box lines with texts being noise. Recall in original DeAE, authors recommended noisy inputs should not drift too much from  original data distribution. In Line-DDeCAE, we evidently let noise be really different from original manifold. However, in experiments, we find this method works quite well. We assume that though noise(texts) is far from original manifold(box lines), box lines lie in more compact manifold and recovering it from higher manifold is relatively easy. Finally, we add original input image with Line-DDeCAE's output in order to get clean images without box lines. 

In conclusion, suppose the original input with box lines is $\upsilon $, the Line-DDeCAE's output is $l$, then clean image is $\widetilde{\upsilon }$, the above procedure can be summed up in the following operations:

\begin{align}
l&=Line\_DDeCAE(\upsilon) \\
\widetilde{\upsilon }&=l+\upsilon
\end{align}

\subsection{Knowledge Distillation}
In this section, we introduce one simple KD method for compressing large(teacher) CRNN model into a small(student) one.

Standard KD uses KL divergence to push student's softmax to be close to teacher's with temperature hyperpatameter $T$, which is called soft target. Concretely, standard KD applies loss function combined soft targets with original targets(hard labels). Suppose the logit of teacher model is $L_t$ and logit of student model is $L_s$ and $y^{(i)}$ is corresponding ground-truth label, $i^{th}$ example's softmax output of student model is $f(L_s^{(y^{(i)})})=\frac{exp(L_s^{(y^{(i)})})}{\sum_{j=1}^{K}exp(L_s^{(y^{(j)})})}$, and softmax output of teacher model is $f(L_t^{(y^{(i)})})=\frac{exp(L_t^{(y^{(i)})})}{\sum_{j=1}^{K}exp(L_t^{(y^{(j)})})}$, the loss of standard KD $\mathcal{L}_{SKD}$ for student model is: 

\begin{eqnarray}
	\mathcal{L}_{SKD}=
		-\frac{\alpha} {B}\sum_{i=1}^{B}log\left ( f(L_s^{(y^{(i)})}) \right )-\frac{1-\alpha}{B} \sum_{i=1}^{B} \left ( f(L_t^{(y^{(i)})} / T)\right)log\left (f(L_s^{(y^{(i)})} / T) \right )
\end{eqnarray}

where $\alpha$ is hyperparameter to trade off hard targets cross entropy loss and soft targets KL divergence.

\begin{figure}[t]
	\begin{center}
		\centering
		\includegraphics[width=14cm,height=8cm]{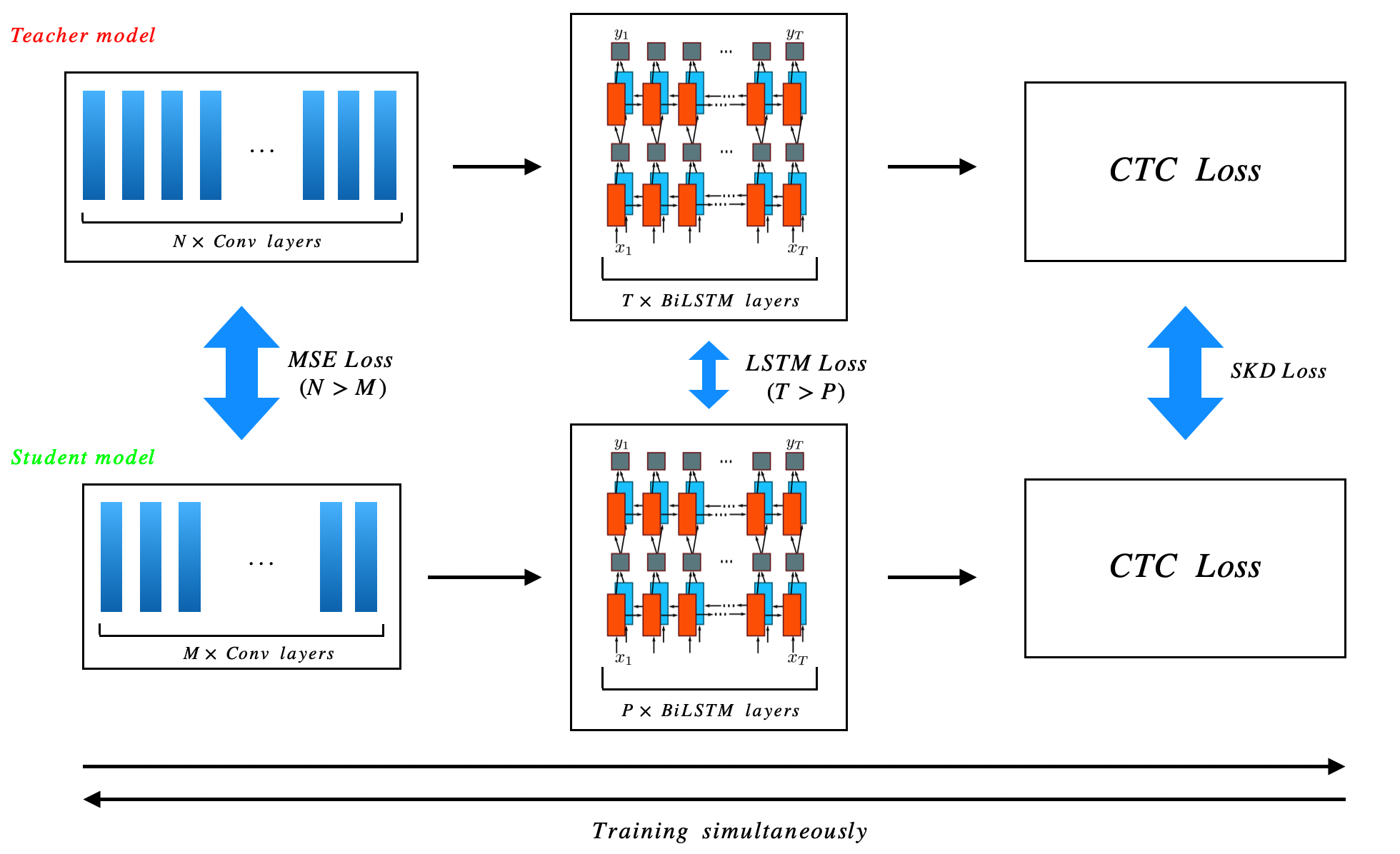}
	\end{center}
	\caption{KD training architecture}\label{kd}
\end{figure}

In addition to standard KD, in CNN we also adopt the idea from FitNets (\cite{romero2014fitnets}) to compare intermediate layers activations between teacher model and student model. In order to match the dimension, we use $1*1$ convolution layer as regressor layer. 
Assume we compare $M$ layers, for $l^{th}$ layer, the activation of student model is $o_{s}^{(l)}$ and the activation of teacher model is $o_{t}^{(l)}$, then all $M$ layers MSE loss $\mathcal{L}_{mse}$ is:
  
\begin{eqnarray}
	\mathcal{L}_{mse} = \frac{1}{MB}\sum_{i=1}^{B}\sum_{l=1}^{M}(regressor(o_{s}^{(i)(l)}) - o_{t}^{(i)(l)})^{2}
\end{eqnarray}

Finally, to distill knowledge from BiLSTM as well, we propose to compare last time step of $c$ values and $h$ values between student model and teacher model because the last step of hidden values in BiLSTM may accumulate enough information to distill. Following the same notations above, denote last $h$ in teacher model as $h_{s}$ and student model as $h_{t}$, we can write $\mathcal{L}_{lstm}$ as:

\begin{equation}
\mathcal{L}_{lstm}=\frac{1}{B}\sum_{i=1}^{B}\left [ (c_{t}^{(i)}-c_{s}^{(i)})^2+(h_{t}^{(i)}-h_{s}^{(i)})^2 \right ]
\end{equation}

We use the same hidden dimensions in both teacher model and student model.

To put it all together, we use the following loss function in optimization:
\begin{equation}
\mathcal{L}_{KD}=\mathcal{L}_{SKD}+\lambda_1\mathcal{L}_{mse}+\lambda_2\mathcal{L}_{lstm}
\end{equation}

One thing to note that, we optimize the whole student model along with teacher model simultaneously, to overcome model collapse problem (\cite{miyato2018virtual}), we detach value of teacher model from PyTorch tensor to only encourage student model to be optimized towards distilled manner. Therefore the whole optimized loss function is :
\begin{equation}
\mathcal{L}=\mathcal{L}_{ctc}+\mathcal{L}_{KD}
\end{equation}

As a whole, the KD algorithm proposed in this paper can be seen in Figure \ref{kd}.

\section{Experiments}
\label{experiments}
\subsection{Data Sets}
 In this paper, we generate simple synthetic data sets from a Chinese novel book: SanTi. We generate all samples following the order of the book. Each image includes 1 row or 7 rows or 10 rows and similar columns to make each image a near square shape except for single-row images. We create single-row images data set, called $S_{data1}$ for CAP verification. 
 For MultiRow Stretching and Attention methods, we generate $M_{data1}$(each image contains 7 rows)  without box lines, and we also create multi-row images with box lines data set $M_{data2}$(each image contains 10 rows) to verify Line-DDeCAE method. For KD, we run on extended version of $S_{data1}$ which is $S_{data2}$ and $M_{data1}$ for distillation.
 
\begin{figure}[h]
 	\centering
 	\subfigure[$S_{data1}$ and $S_{data2}$]{
 		\centering
 		\includegraphics[width=2.7cm]{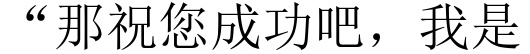}
 	}%
 	\subfigure[$M_{data1}$]{
 		\centering
 		\includegraphics[width=2.7cm]{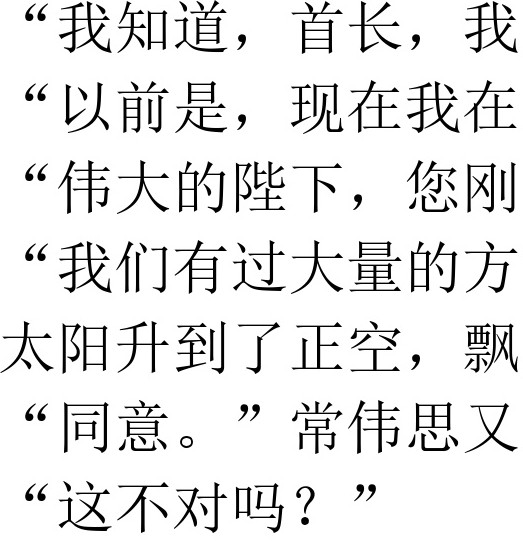}
 	}%
 	\subfigure[$M_{data2}$]{
 		\centering
 		\includegraphics[width=2.7cm]{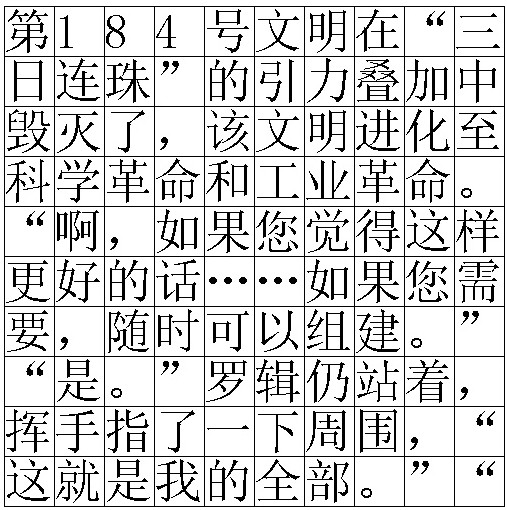}
 	}%
 	
 	\centering
 	\caption{Data sets samples}\label{data_sets}
 \end{figure}

 For each method, we split corresponding data set into 80\% for training and 20\% for testing. Owing to the ease of CAP algorithm, we only generate 5,000 examples for training and testing. For MultiRow Stretching and Attention methods, we generate 30,000 examples overall. In KD, because we need to train teacher and student models at the same time, we ought to enlarge $S_{data1}$ to 10,000 examples $S_{data2}$ with more characters in each image. Table \ref{dataset} summarizes statistics of the data sets for different methods and Figure \ref{data_sets} gives sampled examples from data sets to illustrate data set styles.
 
 \begin{table}[t]
 	\caption{Data sets statistics}
 	\label{dataset}
 	\begin{center}
 	\begin{tabular}{cccccc}
 		\hline
 		data set  & verified methods              & \# examples & \# train & \# test  & \# rows \\ \hline
 		$S_{data1}$  & CAP                           & 5,000       & 4,000    & 1,000 & 1  \\
 		$S_{data2}$ & KD                            & 10,000      & 8,000    & 2,000  & 1 \\
 		$M_{data1}$ & MultiRow Stretching/Attention/KD & 30,000      & 24,000   & 6,000  & 7 \\
 		$M_{data2}$ & Line-DDeCAE                   & 30,000      & 24,000   & 6,000  & 10 \\ \hline
 	\end{tabular}
\end{center}
 \end{table}
 
 \subsection{Implementation}
 During training, we resize the input image into $64*128$ for single-row images and $224*224$ for multi-row images, we also turn all RGB images into gray scale. In CNN module, we follow the same hyperparameters setting in original VGG16 model but cut a few layers off to avoid overfitting except in the last pooling layer, we use pool size of (2, 3) for single-row images. In BiLSTM module, we set hidden size to be 256 and stack 2 BiLSTM layers. During training, for most of experiments, we set training epochs to be 200 with batch size 16, we try SGD optimizer with Nesterov mode with momentum 0.9 and we also try Adam with AMSGrad mode with momentum 0.9 and RMSProp 0.99, but we only report superior SGD results in single-row images and superior Adam results in multi-row images. Initial learning rate is set $1e-4$ with decay rate of 2 for every 60 epochs. We also set weight decay parameter to be $1e-6$. In KD, $\mathcal{L}_{mse}$ with parameter being 0.5 and $\mathcal{L}_{lstm}$ with parameter being 0.2, we set temperature to be 1.5, we obtain above hyperparameters with grid search by spliting training data into 90\% training and 10\% validation. For student model, we cut off a few layers in CNN module and only use one layer of BiLSTM. More details with hyperparameters can be found in Appendix \ref{training}.
 
 \subsection{Metrics}
 In this paper, we use two metrics: character-level precision(CLP) and image-level precision(ILP). In CLP, we count predicted characters in each image against all characters, in ILP, if all characters in each image are predicted correctly, we get one right result. Concretely, 
 
 \begin{eqnarray}
 CLP=\frac{\sum_{i=1}^{m}\sum_{j=1}^{n^{(i)}}\mathbb{I} \left ( y_{j}^{(i)} = \widetilde{y}_{j}^{(i)}\right )  }{\sum_{i=1}^{m}n^{(i)}} \\
 ILP=\frac{\sum_{i=1}^{m}\mathbb{I} \left ( y^{(i)}=\widetilde{y}^{(i)} \right ) }{m}
 \end{eqnarray}
 
 where $m$ is the number of examples and $n^{(i)}$ is the number of characters in $i^{th}$ example, $y$ is ground-truth and $\widetilde{y}$ is the corresponding prediction, $\mathbb{I}$ is indicator function.
 
 \subsection{Results}
 We report all averaged results on test set with three runs each. 
 \subsubsection{CAP}
 In this part, we use $S_{data1}$ to compare CAP with original CRNN model. In table \ref{cap_res}, as comparison, we use the default setting in CRNN model. Notice CAP performs almost equally to original CRNN model both in CLP(-0.09\%) and ILP(-0.08\%). We argue that CAP can also obtain similar results without hand tuning hyperparameters in CNN module.
 
 \begin{table}[h]
 	\caption{Comparison between CAP and CRNN, all results are reported with \%.}
 	\label{cap_res}
 	\begin{center}
 	\begin{tabular}{ccc}
 		\hline
 		methods & CLP   & ILP   \\ \hline
 		CRNN    & \textbf{97.31} & \textbf{91.33} \\
 		CAP     & 97.20 & 91.25 \\ \hline
 	\end{tabular}
 	\end{center}
 \end{table}
 
 \subsubsection{MultiRow Stretching and Attention}
 We use $M_{data1}$ without box lines to verify both MultiRow Stretching and Attention methods. Final results are in Table \ref{multi_res}. Original CRNN can't handle multi-row images, hence we only report two proposed methods. Note that for MultiRow Stretching final results are pretty good compared to Attention method in CLP(+0.76\%), we argue the data set contains only 7-row images, we claim this performance may decline if each image contains more rows. 
 \begin{table}[h]
 	\caption{Comparison between MultiRow Stretching and Attention, all results are reported with \%.}
 	\label{multi_res}
 	\begin{center}
 	\begin{tabular}{ccc}
 		\hline
 		methods             & CLP   & ILP   \\ \hline
 		MultiRow Stretching & 80.52 & \textbf{60.17} \\
 		Attention           & \textbf{81.31} & 60.04 \\ \hline
 	\end{tabular}
 \end{center}
 \end{table}
 
 \subsubsection{Line-DDeCAE}
 Next, we train Line-DDeCAE model using $M_{data2}$. The sampled images are in Figure \ref{res_ddecae}, more samples can be found in the Appendix \ref{more_results}. We also compared with simple DDeCAE model. Line-DDeCAE and DDeCAE share the same model structure. Note that, DDeCAE model fails to recover texts well, however, Line-DDeCAE recovers box lines almost perfectly. We also conclude that Line-DDeCAE also converges faster and better than DDeCAE, training procedures can be found in Appendix \ref{procedure}. Then, to verify more Line-DDeCAE model, we input $M_{data2}$ data into pre-trained Line-DDeCAE model and inject output into CRNN model to train MultiRow Stretching algorithm. All results are in Table \ref{line_res}.  Compared to DDeCAE we notice that, using Line-DDeCAE can obtain high performance both in CLP(+5.09\%) and ILP(+9.51\%). Also notice that if we directly input $M_{data2}$ into CRNN, the model fails to converge at certain level. 
 
 \begin{figure}[h]
 	\begin{center}
 		\centering
 		\includegraphics[width=10cm,height=3.5cm]{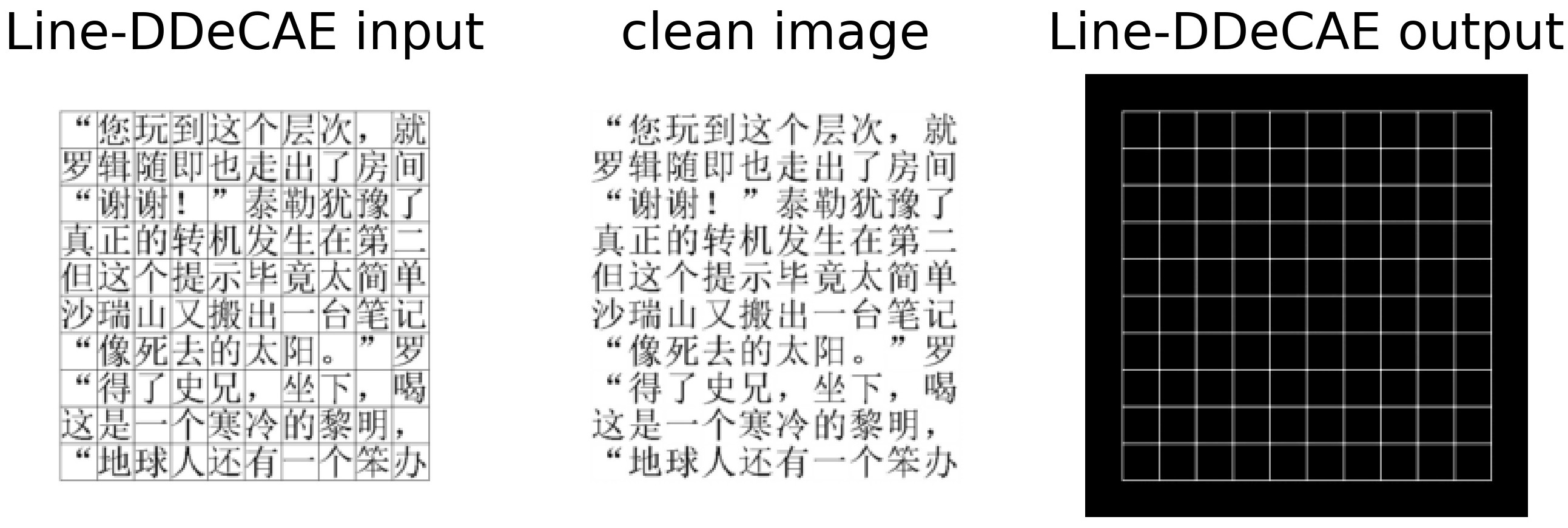}
 	\end{center}
 	\caption{Line-DDeCAE result}\label{res_ddecae}
 \end{figure}
 
 \begin{table}[h]
 	\begin{center}
 		\caption{Comparison between CRNN, DDeCAE and Line-DDeCAE, all results are reported with \%.}
 		\label{line_res}
 	\begin{tabular}{ccc}
 		\hline
 		methods          & CLP            & ILP            \\ \hline
 		CRNN             & 50.77          & 35.98          \\
 		DDeCAE+CRNN      & 70.38          & 50.14          \\
 		Line-DDeCAE+CRNN & \textbf{75.47} & \textbf{59.65} \\ \hline
 	\end{tabular}
 \end{center}
 \end{table}
 
 \subsubsection{KD}
 Finally, we use $S_{data2}$ and $M_{data1}$ to distill knowledge from previous teacher CRNN model into \textbf{half-sized} student CRNN model. To distill knowledge from teacher model efficiently, we train teacher and student model simultaneously. However, we detach values from tensors of teacher model to avoid model collapse problem (\cite{miyato2018virtual}). Final results can be found in Table \ref{distill_res}. In single-row images, performance was improved by a huge gap(+2.45\% in CLP, +7.03\% in ILP) in distilled model, and in multi-row images, we run MultiRow Stretching method, distilled model still outperforms teacher model(+0.22\% in CLP, +0.23\% in ILP). In summary, almost half of parameters are reduced, but similar accuracy is retained and even boosted, which is a very promising result.
 
 \begin{table}[h]
 	\caption{Comparison between CRNN and DistilledCRNN on two data sets, all results are reported with \%.}
 	\label{distill_res}
 	\begin{center}
 	\begin{tabular}{cccc}
 		\hline
 		\multicolumn{1}{l}{data sets} & methods        & CLP            & ILP            \\ \hline
 		\multirow{2}{*}{$S_{data2}$}  & CRNN           & 89.29          & 71.17          \\
 		& Distilled CRNN & \textbf{91.74} & \textbf{78.20} \\ \hline
 		\multirow{2}{*}{$M_{data1}$}  & CRNN(MultiRow   Stretching)           & 80.52          & 60.17          \\
 		& Distilled CRNN & \textbf{80.74} & \textbf{60.40} \\ \hline
 	\end{tabular}
 \end{center}
 \end{table}

\section{Conclusion}
\label{conclusion}
In this paper, we've proposed multiple simple yet effective methods to boost CRNN performance via implementing Chinese text images recognition task. In particular, we advocated to use CAP to replace exhaustive hyperparameters fine-tuning with single-row images; we also proposed two methods to tackle with multi-row images; original CRNN can't recognize excel-like images well, we thus proposed Line-DDeCAE to first recover box lines, which we found is an easy task, then add original input image to obtain clean image; finally, we proposed efficient online KD method to distill teacher model into half-sized student model without loss of generality. Experimental results on generated synthetic data set revealed our methods being of high efficiency.

\newpage
\bibliography{iclr2021_conference}
\bibliographystyle{iclr2021_conference}

\appendix
\newpage
\section{Appendix}
\subsection{Training Hyperparameters}
\label{training}
In this section, we list all hyperparameters in all experiments in Table \ref{hypers}.
\begin{table}[h]
	\caption{All hyperparameters in different training methods.}
	\label{hypers}
	\setlength{\tabcolsep}{0.1mm}
	\begin{center}
	\begin{tabular}{cccccc}
		\hline
		\multirow{2}{*}{methods} & \multicolumn{5}{c}{hyperparameters}                                               \\
		& image size           & epochs       & batch size & optimizer        & init\_lr    \\ \hline
		CRNN                     & (224, 224)/(64, 128) & 200          & 20         & AMSGrad/Nesterov & 1e-4        \\
		CAP                      & (64, 128)            & 200          & 20         & Nesterov         & 1e-4        \\
		MultiRow Stretching      & (224, 224)           & 200          & 20         & AMSGrad          & 1e-4        \\
		Attention                & (224, 224)           & 200          & 20         & AMSGrad          & 1e-4        \\
		DDeCAE                   & (224, 224)           & 100          & 64         & Adam             & 1e-3        \\
		Line-DDeCAE              & (224, 224)           & 100          & 64         & Adam             & 1e-3        \\
		DDeCAE + CRNN            & (224, 224)           & 200          & 16         & AMSGrad          & 1e-4        \\
		Line-DDeCAE + CRNN       & (224, 224)           & 200          & 16         & AMSGrad          & 1e-4        \\
		KD with single-row       & (64, 128)            & 300          & 16         & Nesterov         & 1e-4        \\
		KD with multi-row        & (224, 224)           & 300          & 16         & AMSGrad          & 1e-4        \\ \hline
		& lr\_decay rate       & weight decay & $\alpha$   & $\beta$          & temperature \\ \hline
		CRNN                     & 2                    & 1e-6         & N/A        & N/A              & N/A         \\
		CAP                      & 2                    & 1e-6         & N/A        & N/A              & N/A         \\
		MultiRow Stretching      & 2                    & 1e-6         & N/A        & N/A              & N/A         \\
		Attention                & 2                    & 1e-6         & N/A        & N/A              & N/A         \\
		DDeCAE                   & 5                    & 0            & N/A        & N/A              & N/A         \\
		Line-DDeCAE              & 5                    & 0            & N/A        & N/A              & N/A         \\
		DDeCAE + CRNN            & 2                    & 1e-6         & N/A        & N/A              & N/A         \\
		Line-DDeCAE + CRNN       & 2                    & 1e-6         & N/A        & N/A              & N/A         \\
		KD with single-row       & 2                    & 1e-6         & 0.5        & 0.2              & 1.5         \\
		KD with multi-row        & 2                    & 1e-6         & 0.5        & 0.2              & 1.5         \\ \hline
	\end{tabular}
\end{center}
\end{table}

In the table, init\_lr means initial learning rate, lr\_rate means learning rate decay rate, $\alpha$ and $\beta$ are hyperparameters to trade off between $\mathcal{L}_{mse}$ and $\mathcal{L}_{lstm}$.

\subsection{Training Procedures in Line-DDeCAE and DDeCAE}
\label{procedure}
\begin{figure}[h]
	\centering
	\subfigure[Line-DDeCAE training procedure]{
		\centering
		\includegraphics[width=5cm]{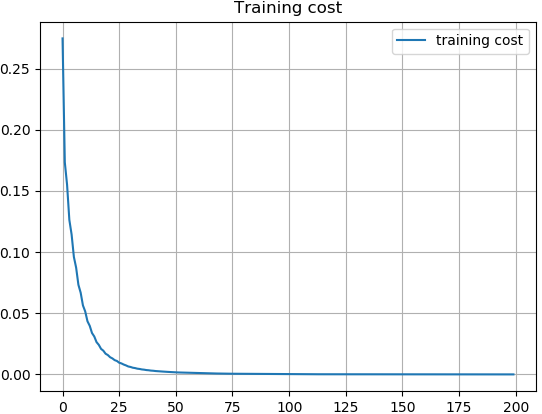}
	}%
	\quad
	\subfigure[DDeCAE training procedure]{
		\centering
		\includegraphics[width=5cm]{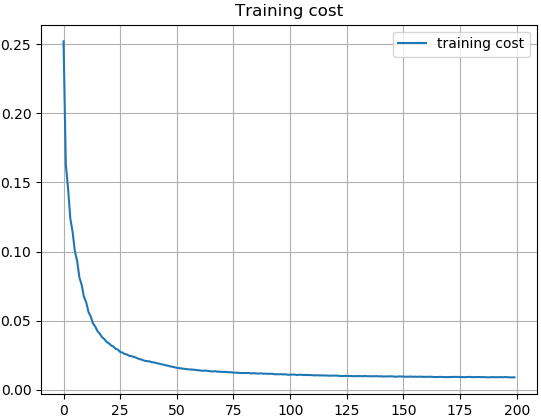}
	}%
	
	\centering
	\caption{Training procedures}\label{train_details}
	
\end{figure}

Figure \ref{train_details} reveals how training losses change both in Line-DDeCAE and DDeCAE, we can see Line-DDeCAE converges slightly faster and much better than DDeCAE, we argue that since box lines are easily to be recovered in high dimensional space than texts, therefore, Line-DDeCAE is a reasonable alternative in this scenario. 

\subsection{More Samples from Line-DDeCAE and DDeCAE}
\label{more_results}
\begin{figure}[h]
	\centering
	\subfigure{
		\centering
		\includegraphics[width=12.7cm]{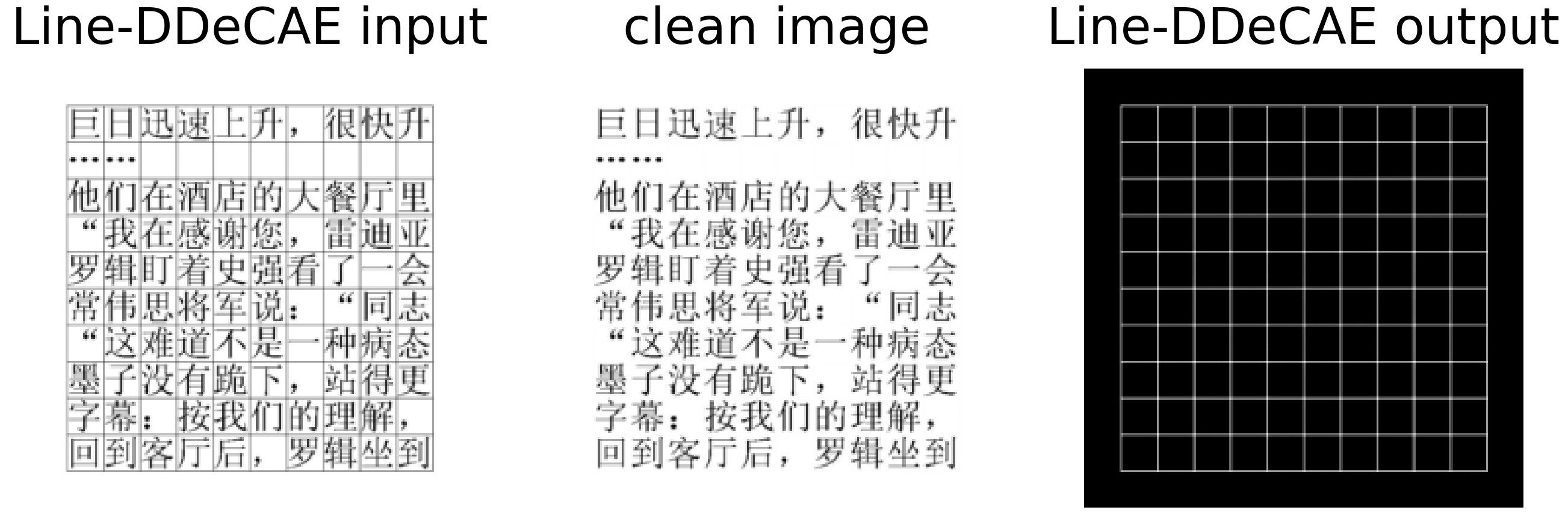}
	}%
	\quad
	\subfigure{
			\centering
			\includegraphics[width=12.7cm]{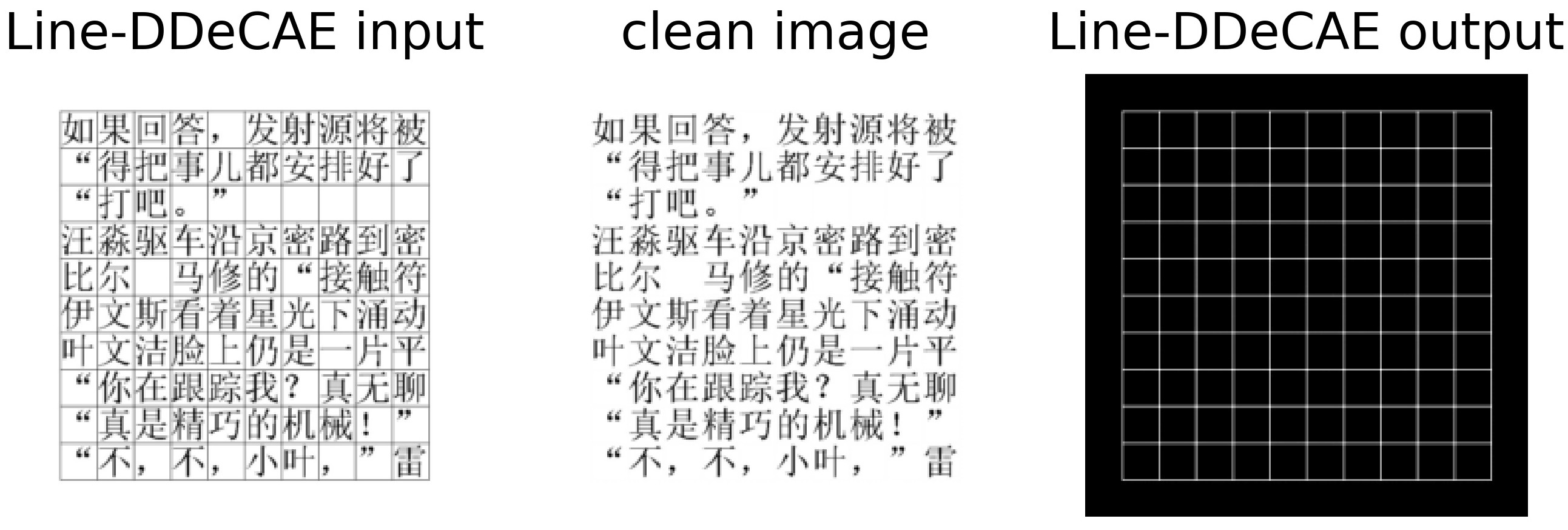}
	}%

	\subfigure{
			\centering
			\includegraphics[width=12.7cm]{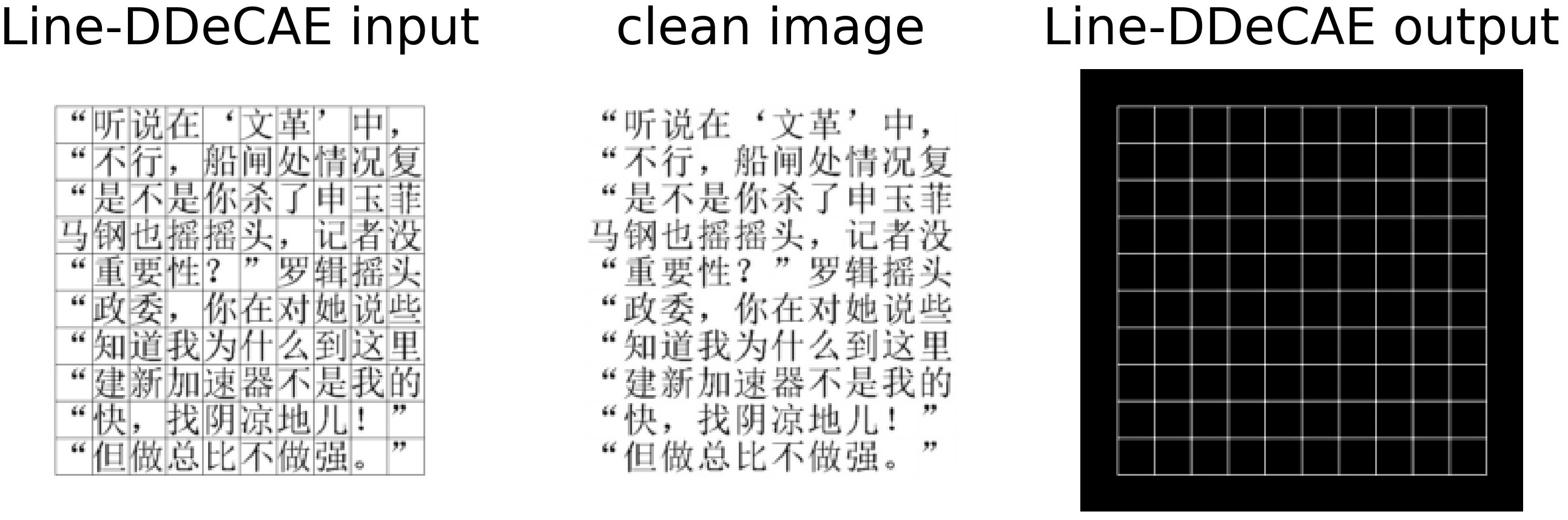}
	}%

	\subfigure{
			\centering
			\includegraphics[width=12.7cm]{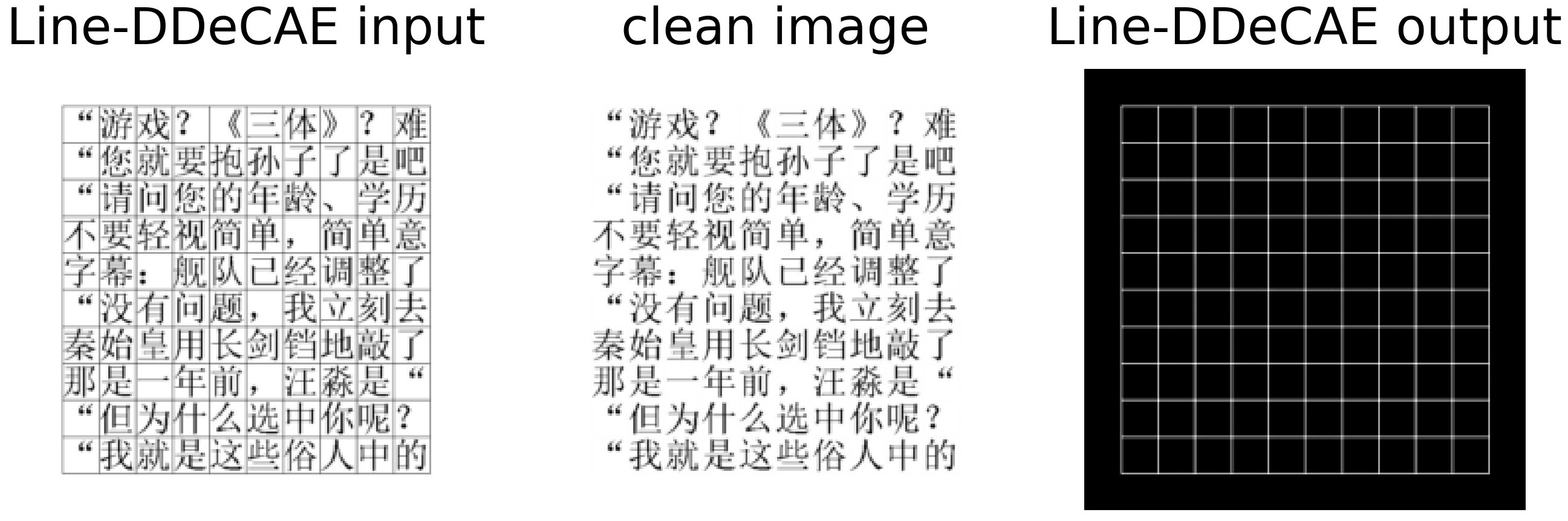}
	}%

	\centering
	\caption{Line-DDeCAE results}\label{more_lineddecae}

\end{figure}

In this section, we display more results on Line-DDeCAE and DDeCAE for comparison in Figure \ref{more_lineddecae} and Figure \ref{more_ddecae} respectively.

\begin{figure}[t]
	\label{more_ddecae}
	\centering
	\subfigure{
		\centering
		\includegraphics[width=8.3cm]{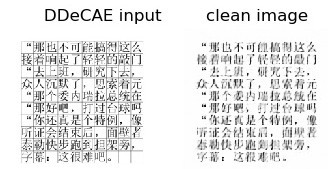}
	}%

	\subfigure{
		\centering
		\includegraphics[width=8.3cm]{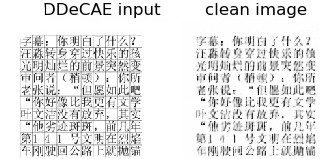}
	}%
	
	\subfigure{
		\centering
		\includegraphics[width=8.3cm]{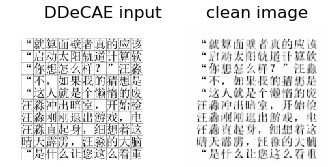}
	}%

	\subfigure{
		\centering
		\includegraphics[width=8.3cm]{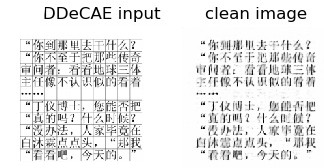}
	}%
	
	\subfigure{
		\centering
		\includegraphics[width=8.3cm]{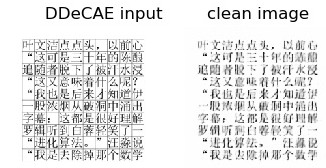}
	}%
	\centering
	\caption{DDeCAE results}\label{more_ddecae}
	
\end{figure}

According to figures, we conclude that Line-DDeCAE explicitly recovers box lines perfectly, whereas DDeCAE tries harder to recover texts which is not ideal.

\end{document}

%% file: math_commands.tex
\usepackage{amsmath,amsfonts,bm}

\def\eqref#1{equation~\ref{#1}}

\def\1{\bm{1}}

\DeclareMathAlphabet{\mathsfit}{\encodingdefault}{\sfdefault}{m}{sl}
\SetMathAlphabet{\mathsfit}{bold}{\encodingdefault}{\sfdefault}{bx}{n}

